\documentclass[conference]{IEEEtran}
\IEEEoverridecommandlockouts
\usepackage{cite}
\usepackage{amsmath,amssymb,amsfonts}
\usepackage{algorithmic}
\usepackage{graphicx}
\usepackage{textcomp}
\usepackage{xcolor}
\usepackage{todonotes}
\usepackage{verbatim}
\usepackage{subfig}
\usepackage{url} 

\def\BibTeX{{\rm B\kern-.05em{\sc i\kern-.025em b}\kern-.08em
    T\kern-.1667em\lower.7ex\hbox{E}\kern-.125emX}}
\begin{document}

\title{Perfecting Aircraft Maneuvers with\\ Reinforcement Learning
}

\makeatletter
\newcommand{\linebreakand}{%
  \end{@IEEEauthorhalign}
  \hfill\mbox{}\par
  \mbox{}\hfill\begin{@IEEEauthorhalign}
}
\makeatother

\author{\IEEEauthorblockN{Atahan Cilan}
\IEEEauthorblockA{
\textit{Turkish Aerospace}\\
Istanbul, Turkey \\
atahan.cilan@tai.com.tr}
\and
\IEEEauthorblockN{Mahir Demir}
\IEEEauthorblockA{
\textit{Turkish Aerospace}\\
Istanbul, Turkey \\
mahir.demir@tai.com.tr}
\and
\IEEEauthorblockN{Ozgun Can Yurutken}
\IEEEauthorblockA{
\textit{Turkish Aerospace}\\
Istanbul, Turkey \\
ozguncan.yurutken@tai.com.tr}
\and
\IEEEauthorblockN{Seyyid Osman Sevgili}
\IEEEauthorblockA{
\textit{Turkish Aerospace}\\
Istanbul, Turkey \\
seyyidosman.sevgili@tai.com.tr}
\linebreakand
\IEEEauthorblockN{Umit Can Bekar}
\IEEEauthorblockA{
\textit{Turkish Aerospace}\\
Istanbul, Turkey \\
can.bekar@tai.com.tr}
}

\maketitle

\begin{abstract}
This paper evaluates an advanced jet trainer's utilization of artificial intelligence (AI)-based aircraft aerobatic maneuvers with the intention of developing an AI-assisted pilot training module for specific aircraft maneuvers. A multitude of aircraft maneuvers have been simulated using reinforcement learning (RL) agents, which will serve as a training tool for future pilots. 
\end{abstract}


\section{Introduction}
We have developed two distinct methods for performing maneuvers with distinct use cases. However, the results demonstrated that AI is capable of learning any intended maneuver, irrespective of the method employed.

Both methods utilize a sophisticated RL foundation, incorporating cutting-edge soft-actor critic (SAC) models and hyper-parameter optimization methods. The algorithm considers classical state variables including roll, gamma, yaw, and speed errors of the aircraft, as well as angles of control surfaces, attitude, speed, and altitude. 

The initial approach involves utilizing the flight records of the example pilot as the input and designating the targets (roll, gamma, yaw, speed) as the subsequent states specified in the flight records. Using an appropriate reward function, the AI model was able to acquire the ability to replicate the actions of the exemplar pilot. However, in order for this method to function properly, the pilots' example must precisely match their expectations, as the AI model will replicate their actions. The second approach is predicated on the condition stated previously. If the pilot is unable to provide the desired input data for the AI model to accurately imitate, the second method can be used to train the model. The second approach involves constructing and inputting the mathematical model of the intended maneuver, including the target yaw, roll, pitch degrees, and speed at each time step, rather than simulating the flight data.

In both of these approaches, the reward/punishment paradigm is utilized. Examining the paradigm consists of three components. As the model's primary states (roll, gamma, yaw, speed) approach the primary target, it ought to be rewarded initially. Furthermore, in a case that the aircraft deviates significantly from the intended value, the episode can be terminated. As a final requirement, the aircraft must execute the target maneuvers neither oscillating nor vibrating; therefore, an additional penalty has been imposed for such conduct. Command inputs of the three control surfaces comprise the action space: the aileron, elevator, rudder, along with the throttle command.

Our final models, which have been evaluated using simulation data and test pilots, can match the quality of a professional pilot, which demonstrates that the capabilities of artificial intelligence can be applied to educational scenarios.

In summary, our research equips an advanced jet training simulator with maneuvering capabilities by implementing sophisticated RL methodologies. A contemporary reinterpretation of RL baseline and aerospace engineering domain-specific reward/punishment systems is employed to achieve this. 

Our main contributions and findings are listed as follows:

\begin{itemize}
\item \textbf{Generality of the pipeline:} The proposed algorithm can be used for any physically acceptable maneuver if the trajectory of the maneuver is available (can be real pilot data or an artificially created trajectory).
\item \textbf{Real pilot data usage:} We utilized real pilot data in our studies thanks to Turkish Aerospace, therefore we showed that AI solutions can match the quality of a professional pilot. Most of the studies use data generated by volunteers who have little to no real flight experience \cite{andrew2018reinforcement, medeiros2021learn,de2022learn}.
\item \textbf{Maneuver stability:} Supervised learning-based solutions for aerobatic maneuvers generally have stability problems \cite{medeiros2021learn}. We eliminated these problems by utilizing RL and training the models on a wide range of initial conditions.
\item \textbf{Maneuver scalability:} We extended our algorithm to demonstrate the possibility of generating scaled versions of a maneuver by using the same reference trajectory and revealed the constraints of the related method.
\item \textbf{Noise reduction with domain knowledge:} We demonstrated the possibility of creating noise-free maneuver trajectories by using domain knowledge, which led to more precise maneuvers when compared to real pilot maneuvers.
\item We have seen that longer maneuvers require more training time and are inclined to accommodate errors, which implies the necessity for short and inclusive maneuvers. Also, it is possible to generate complex maneuvers by consecutively executing different maneuvers via trained models, which is an indicator of the stability of our models.
\end{itemize}

The rest of the paper is organized as follows:

In Section \ref{pw}, a short history of previous works related to AI applications on aerobatic maneuvers is given. In Section \ref{bg}, general structure of the RL algorithms are mentioned. In Section \ref{method}, methodology of the paper is explained. Section \ref{result} discusses the results and Section \ref{conclusion} concludes the paper. 

\section{Previous Works}\label{pw}
Aerobatic maneuvers and reinforcement learning are two distinct yet cooperative areas. Due to the gamification ability of reinforcement learning, in aircraft literature, it was used in dogfight-like combat scenarios like ones in \cite{yuan2022research,chai2023hierarchical,li2024cross,jia2023multi}. 

On the other hand, it is somehow complex to set an RL scheme, especially for the sophisticated area of aircraft. Thus, supervised learning methods are preferable if one can get data on air vehicle flight. 

The most prominent research in this area is the paper by Abbeel et al.\cite{abbeel2006application}. In this paper, the authors used their self-made helicopter to first generate trajectories of the intended maneuver. Afterward, they design a controller based on apprenticeship learning methodologies. We have used similar logic in our work for our vehicle.

In \cite{medeiros2021learn}, one common aerobatic maneuver, Immelmann Turn was implemented. The authors used behavioral cloning methodology, i.e. providing a human-generated trajectory and cloning the same trajectory with AI-based methodology. They have used 30 Immelmann Turn demonstrations with MSE loss, for both ANN and LSTM network structures.\cite{freitas2023performing} extends the work of \cite{medeiros2021learn} to other aircrafts and maneuvers. The authors still used LSTM to take the flight data from normal users from FSX. As the data does not come from the expert pilots, it has the weakness of not being realistic to use in real-world applications.

In\cite{clarke2020deep}, a Q-learning-based learning methodology was depicted. The authors inspected level flight, slow roll, and knife-edge maneuvers in detail, generated their necessary state values of the trajectory for each time step and shaped the rewards w.r.t generated values. 

\cite{sever2023integrated} provides an alternative approach for the case where pilot demonstrations are not available to train upon. For the initial case, they developed an LSTM-based imitation learning approach, improving the Dagger algorithm by adding a confidence threshold to it for a general airplane. Afterward, they used the transfer learning approach to transfer the learned imitation model to other airplanes. They also provide RL-based approach to update the learned model if aircraft design has changed.

\section{Background}\label{bg}
As we have mentioned previously, we are using the RL paradigm to learn aerobatic maneuvers for a desired aircraft. Thus, this section aims to provide a short overview of RL. 

Reinforcement learning is one of the three main learning methods of machine learning, the others being supervised learning and unsupervised learning. Reinforcement learning is explained as learning what to do -how to match the current state to the next action- so that that action maximizes a specific numerical reward. One more important difference is the actions to take are not given to the agent, like most of the other machine learning methods, thus it must learn them via trial and error. \cite{andrew2018reinforcement}. 

In RL, the agent, which is a mapping between observation and action space (generally referred as policy), observes a state $s_{t}$ from its environment at time step t. Then the agent decides on an action $a_{t}$. The action acts upon the environment and the total system shifts into the state $s_{t+1}$. The action will cause the environment to generate reward $R_{t+1}$\cite{arulkumaran2017brief}, which is the main feedback used for learning in RL. Generated data (state, action, reward pairs) is used to improve the current policy.

The below figure depicts a simple working principle of RL:
\begin{figure}[htbp]
\centerline{\includegraphics[width=0.5\textwidth]{rl-algo.png}}
\caption{Inner working of RL}
\label{fig}
\end{figure}

\section{Methodology}\label{method}
In this study, we used a supervised learning-like approach to construct our objective (or reward) function. As discussed in section \ref{pw}, in the study of Abbeel et al., pilots have demonstrated maneuvers several times, and an algorithm is utilized for generating the intended trajectory of the maneuver. Then, a linear quadratic regulator (LQR) based controller is obtained \cite{abbeel2006application}. By following a similar approach, we utilized the trajectories that are generated by test pilots. Since quadcopters are used in the study of Abbeel et al., the cost of obtaining trajectories is relatively low when compared to our scenario. In our case, maneuver trajectories are generated by experienced pilots in a sophisticated simulator environment, which results in a higher cost per maneuver trajectory. However, generated trajectories are much closer to the intended trajectory for executing maneuvers due to the capability of professional pilots. In this light, we obtained a single trajectory for each maneuver and used them as a reference for our reinforcement learning approach. Furthermore, after analyzing the generated trajectories, we noticed that a recognizable structure exists for most of the maneuvers consisting variable amount of noise in it. As an example, the loop maneuver starts with wings-level flight (gamma and roll angles are zero) and continues by linearly increasing the gamma angle to 90 degrees while maintaining zero roll angle and the initial yaw angle. By using this notion, we carefully analyzed maneuvers and created artificial trajectories by using our domain knowledge. As a result of this effort, we obtained noise-free versions of the trajectories, which is mentioned as the second method in the introduction part.

A trajectory consists of 4 states of the aircraft as a time series data. These states are roll, gamma, yaw angles, and Mach value of the aircraft. Gamma angle is used instead of pitch angle by considering that different aircraft have different properties (mass etc.) and they can follow different flight paths while maintaining the same pitch angle. The main idea behind defining trajectories as 4-channel time series data is finding a minimum number of state variables that express the overall maneuver trajectory properly. These variables could be the positions of the aircraft as time-series data, however, this would create problems when applying the same algorithm to different aircraft. Furthermore, a maneuver can create different trajectories at distinct conditions such as varying altitude, initial position, etc if the trajectories are constructed by the positions of the aircraft. By defining a trajectory in terms of attitude and speed, we aimed to capture the essence of the maneuver, expressing a maneuver with the minimum number of state variables.

By utilizing these trajectories, we know the target attitude (orientation) and speed of the aircraft at each time step for a desired maneuver. As the next step, we converted these trajectories to a reward function for using them with a reinforcement learning algorithm. It is convenient to think of a trajectory as a series of target points. An aircraft can be trained to reach a target attitude and speed by defining the reward in terms of the error related to the desired target \cite{pars2024}. We used a similar approach to convert obtained trajectories to a reward function. Instead of a static target, we utilized time-varying targets and calculated reward by using the error for the current target in the trajectory. In Figure \ref{traject-algo}, the trajectory of the loop maneuver with target points is displayed. At each time step, the target point is sampled from the maneuver trajectory, and errors related to these targets are given as input to the RL model. It is possible to use different sampling rates for sampling a target point from the trajectory. If the sampling frequency is equal to the agent's step frequency, which is the number of actions done by the agent in a second, then the agent observes all of the target trajectory. If the sampling frequency is the half of agent's step frequency (green dots in Figure \ref{traject-algo}), the agent observes a reduced version of the target trajectory and this can lead to a less smooth maneuver. Therefore, we utilized the same frequency for both of them if the time scaling was not applied. The time scaling of a maneuver is discussed in the subsection \ref{spaces}. In the end, the objective of the RL algorithm is to find a good policy that can track the time-varying target points. When the model correctly tracks the target points, the overall trajectory of the maneuver is obtained.

As a reinforcement learning algorithm, we selected Soft Actor-Critic (SAC) due to its performance \cite{sac-paper}. We used an altered version of the Gym-JSBSim environment \cite{gym-jsbsim} to train RL models. We actualized maneuvers on both F16 aircraft, which is available in JSBSim, and Hurjet Aircraft, which is available in .dll form and added to the Gym-JSBSim environment by us.

The subsection \ref{lemans} defines the aerobatics maneuvers realized in this study. Observation and action spaces of the reinforcement learning agent are discussed in subsection \ref{spaces}. Subsection \ref{init} explains how we selected the initial conditions when training the model, and subsection \ref{rf} explains the construction of the reward function.

\begin{figure}[htbp]
\centerline{\includegraphics[width=0.5\textwidth]{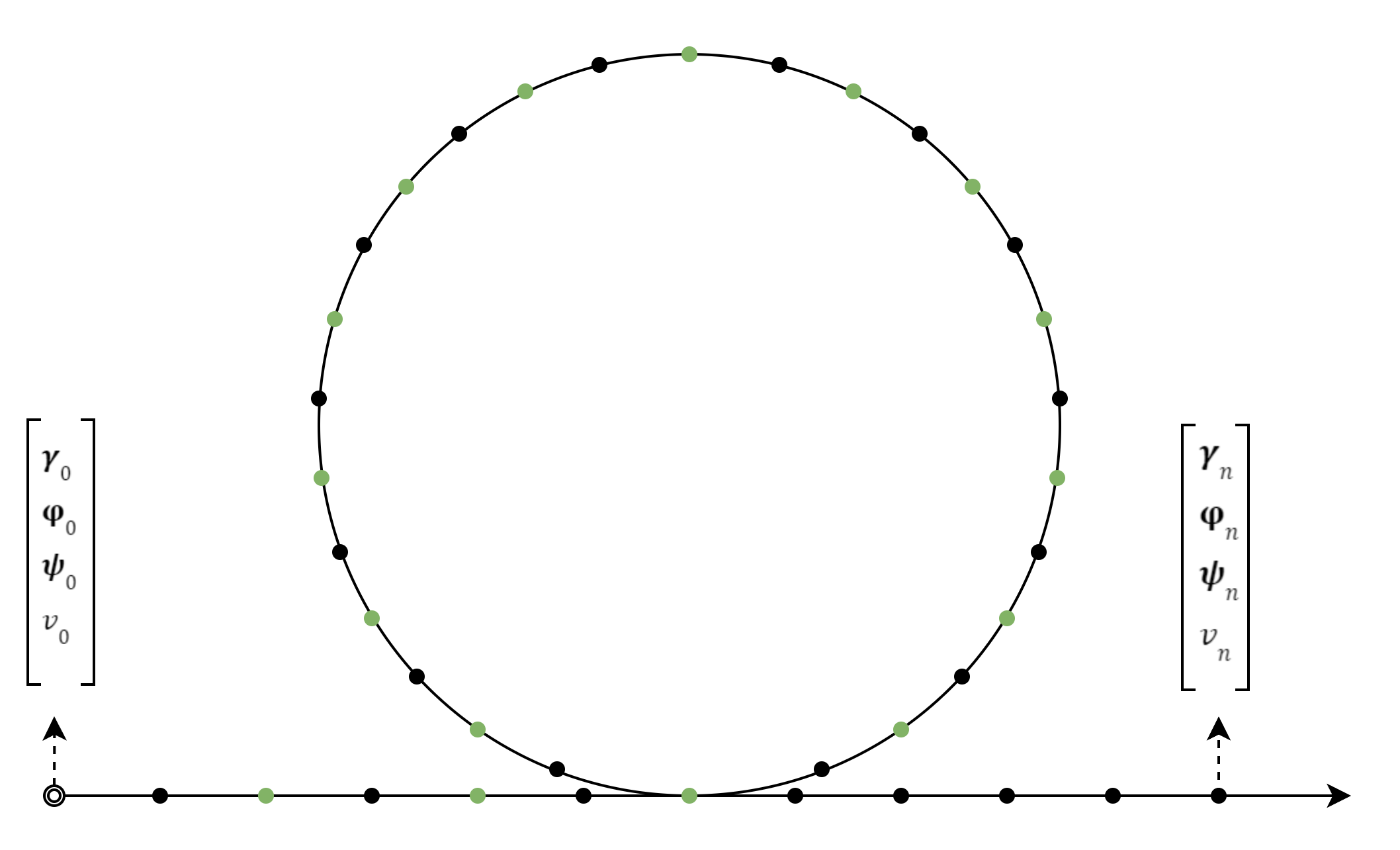}}
\caption{A sample loop maneuver trajectory is displayed. The trajectory is a 4-channel time series data. Each point in the trajectory has 4 components which are gamma, phi, and psi angles, and Mach value. Green dots represent the perceived trajectory with a halved sampling frequency.}
\label{traject-algo}
\end{figure}

\subsection{Maneuvers}\label{lemans}

\subsubsection{Barrel Roll}
A barrel roll is a maneuver where the aircraft makes a complete rotation along its longitudinal axis while also following a helical path, resembling the shape of a barrel. The side view of the barrel roll maneuver is displayed in Figure \ref{barrel-roll-fig}.
\begin{figure}[htbp]
\centerline{\includegraphics[width=0.5\textwidth]{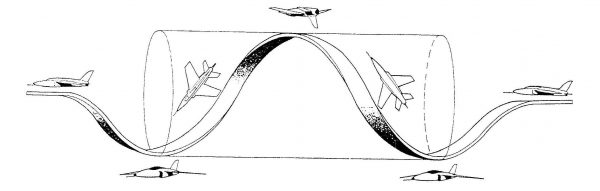}}
\caption{Side view of the barrel roll maneuver \cite{maneuver_fig}.}
\label{barrel-roll-fig}
\end{figure}

\subsubsection{Loop}
A loop maneuver is a 360-degree turn in the vertical axis, which is shown in Figure \ref{loop-fig}. 
\begin{figure}[htbp]
\centerline{\includegraphics[width=0.4\textwidth]{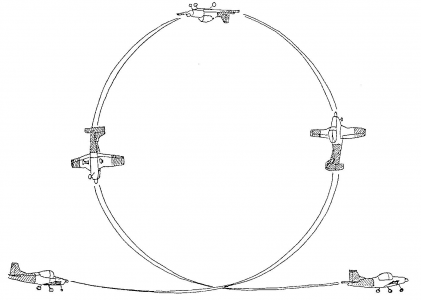}}
\caption{Side view of the loop maneuver \cite{maneuver_fig}.}
\label{loop-fig}
\end{figure}

\subsubsection{Immelmann}
An Immelmann turn is a half-loop maneuver followed by a 180-degree roll, which is shown in Figure \ref{immelmann-fig}.
\begin{figure}[htbp]
\centerline{\includegraphics[width=0.4\textwidth]{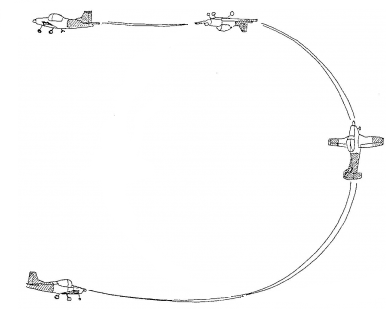}}
\caption{Side view of the Immelmann maneuver \cite{maneuver_fig}.}
\label{immelmann-fig}
\end{figure}

\subsection{Observation and Action Spaces}\label{spaces}
A reinforcement learning agent observes a certain part of an environment and decides on an action based on these observations. Most of the time, the environment is partially observable and the agent has to act with limited knowledge. This is also true in our case. In this study, internal state variables of the aircraft such as altitude, true airspeed, components of velocity vector on each dimension, attitude, previous action, angular rates, and control surface states are included in the observation space along with the errors that will be explained in the subsection \ref{rf}. All of the variables in the observation space are normalized via min-max normalization to the range [0, 1].

Position information (latitude and longitude) is not included in the observation space to prevent possible overfitting to the position. In addition to these variables, a variable called time scaling factor and a remaining maneuver steps variable are added to the observation space. The remaining maneuver steps variable provides information about the agent’s current position on the overall maneuver trajectory. Considering that maneuver targets (the point sampled from the trajectory) change rapidly, introducing this variable enabled the network to associate the remaining step with the successive targets which led to increased learning performance. Since all of the variables are normalized in the end, this variable represents the remaining percentage of the maneuver when it is multiplied by 100, which is independent of the total time steps necessary for the respective maneuver. Various maneuvers require various amounts of time to execute them. As an example, Immelmann is a shorter maneuver than the loop since it is simply a half-loop maneuver followed by a 180-degree roll maneuver. The RL agent, mainly observes the targets for the maneuver and the remaining maneuver percentage, but does not directly perceive the total execution time of the maneuver. Therefore, total execution time of a maneuver can be altered. This notion motivated us to introduce another variable to observation space for extending the maneuver on time. Extending a known maneuver trajectory in time means generating bigger or smaller loops, barrel rolls, etc. Decreasing the rate of sampling targets from maneuver trajectory results in longer maneuvers and the opposite is also true. However, the RL model needs to know this sampling rate to properly execute the desired maneuver on the desired time scale. Otherwise, the same observations can be encountered during different time-scaled versions of a maneuver and the algorithm struggles to converge to an acceptable policy as we have seen in our experiments. Previously mentioned time scale factor is used for this purpose. Assuming a maneuver’s original length is ‘x’ seconds, and the time scaling factor is ‘tau’; the time-scaled maneuver is executed in x * tau seconds. It is important to be careful about the time scaling factor since there are certain physical constraints to execute a maneuver and the applicable time scaling range differs from maneuver to maneuver. Furthermore, training with different time scaling factors is costly and it takes more time for a model to converge. Therefore, we showed the applicability of time scaling on the loop maneuver and kept the time scaling factor constant on the other maneuvers.

The action space of an RL agent defines the variables that the agent can change in the environment. In our study, action space consists of the stick commands that control the angles of control surfaces, which are aileron, elevator, and rudder. Also, the throttle command that controls the engine thrust is included in the action space. The policy network generates actions in the range [-1, 1] for each action space dimension. Then, we denormalize the action to the actual range of stick and throttle commands. The actual range of the action and observation spaces are different for F16 and Hurjet aircraft but we used the same logic for both. Also, it is possible to design the policy network in a way that generates action in the desired range, however, it is mentioned in the Stable Baselines 3 documentation that it is suggested to use symmetric and normalized actions when working with a custom environment \cite{stable-baselines3}.

\begin{figure}[htbp]
\centerline{\includegraphics[width=0.5\textwidth]{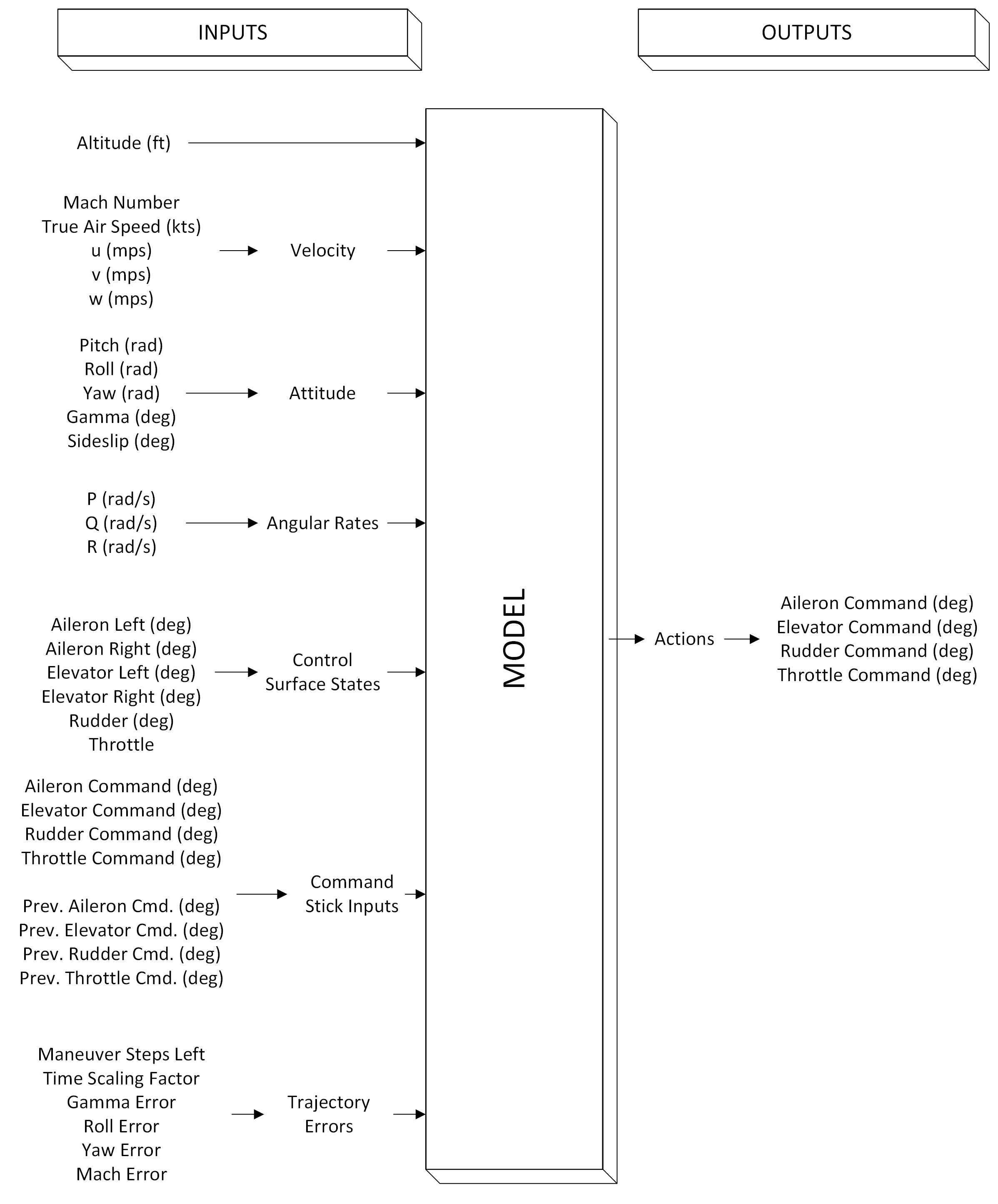}}
\caption{Inputs and outputs of the RL model.}
\label{obs-and-act-space}
\end{figure}

\subsection{Initial Conditions}\label{init}
Determining the initial conditions for training episodes is significant to simulate all possible scenarios for the desired task so that the RL model can have enough state, action, and reward pairs to learn expected behavior. In our work, the aim is to train RL agents that can execute a maneuver in any location and direction, along with the acceptable altitude configurations. Considering that latitude and longitude information is not included in the observation space of the agents, we did not randomize the initial values of these variables. Executing the maneuvers in any direction is crucial and we selected the initial yaw angle randomly from a uniform distribution at the beginning of each training episode. Maneuver trajectories generated by pilots are in a specific direction, so the target points in the trajectory are based on that initial direction. Other targets such as roll and gamma angle, and Mach value are absolute for each maneuver but the yaw angle targets must be relative to the initial yaw angle. Therefore, the initial yaw angle in the trajectory is subtracted from all yaw angles in the trajectory and added to the randomly selected initial yaw angle for each training episode. Similar approach is utilized for artificially created (or handcrafted) trajectories but all of them are constructed by using zero initial yaw angle, therefore substraction part was not necessary.

On the other hand, the initial altitude is kept at 4000 ft since the maneuvers executed by the test pilots are at a similar altitude. The reason for that is most of the maneuvers are aerobatics maneuvers that are executed in the visible range of the audience. In some of the training sessions, we also experimented with variable initial altitudes (4000, 6000, 8000 ft) to validate the generality of our solution. We have seen that the RL agent learns to execute the maneuver at different altitudes, however the training takes longer time. Another thing that we noticed is the agents can execute some of the maneuvers such as loop on different altitudes even though they are not trained with the necessary configuration. Considering that the loop maneuver includes significant changes in the altitude during the maneuver, the agent observes different altitudes during training and can execute the maneuver on different altitudes without further training. In contrast, maneuvers like the lateral loop are mostly executed at the same altitude, and altering the altitude significantly disturbs the maneuver if the model is not trained on various altitude conditions.

\subsection{Reward Function}\label{rf}
Reward function design is one of the challenging processes in reinforcement learning. It provides the necessary information to the reinforcement learning agent about how the desired task should be done. Considering it is only a scalar value at each time step, it should be carefully designed to reflect the desired task's nature properly. In this study, we constructed the reward function by using 8 types of different error components. The first half of them, which are roll, gamma, yaw angle errors, and mach errors, provide information about how close the aircraft is to the desired maneuver trajectory at the current time step. These error types are called asymptotic errors in the Gym-JSBSim framework and errors are normalized by using the following equation:  

\[
e = | x_{\text{target}} - x|,\text{ } e_{\text{normalized}} = \frac{e}{1 + e}
\]

Afterward, normalized errors are subtracted from 1 to obtain the reward components and overall reward is calculated by taking a weighted average of each reward component. The second half of the error components are aileron, elevator, rudder, and throttle command errors. These command errors are simply the difference between current command and previous command.  Applying significantly different commands at consecutive time steps results in oscillations and vibrations. Therefore, we included these error components in the reward function to solve that issue. Contrary to the first 4 error types, these errors are normalized linearly, which means they are scaled to the (0, 1) range by dividing the calculated error by the maximum error. Furthermore, command error components have negative weights to punish rapidly changing commands. 

Both asymptotic errors and linear errors have a scaling factor and a weight factor. The scaling factor is used for normalization, and the weight factor is used in calculating the weighted average. Considering that each error component has a different range, selecting a proper scaling factor is difficult and requires some experimentation. Scaling factors for error components are selected with the help of reward-error graphics, which can be seen in Figure \ref{error-rew-fig}, and experimentation. On the other hand, selecting scaling factors for the command errors is trivial. Since linear normalization is used, the scaling factor is equal to the maximum error range for these components. After determining the scaling factors for each error component, they are kept constant for each maneuver type. Weight factor selection is also significant but relatively intuitive. We utilized different weight factors for each error component and determined them according to their significance for the target maneuver. For instance, the loop maneuver requires precisely controlling the pitch angle (related to the gamma angle in our case), so the weight factor of the gamma angle error component is selected higher relative to other components when training the agent for the loop maneuver. Similarly, maneuvers like barrel roll require better control on the roll axis, and a high weight factor for the roll angle error component is needed. Following this mentality, we created a weight factor set for each maneuver type.

\begin{figure*}
\centerline{\includegraphics[width=0.7\textwidth]{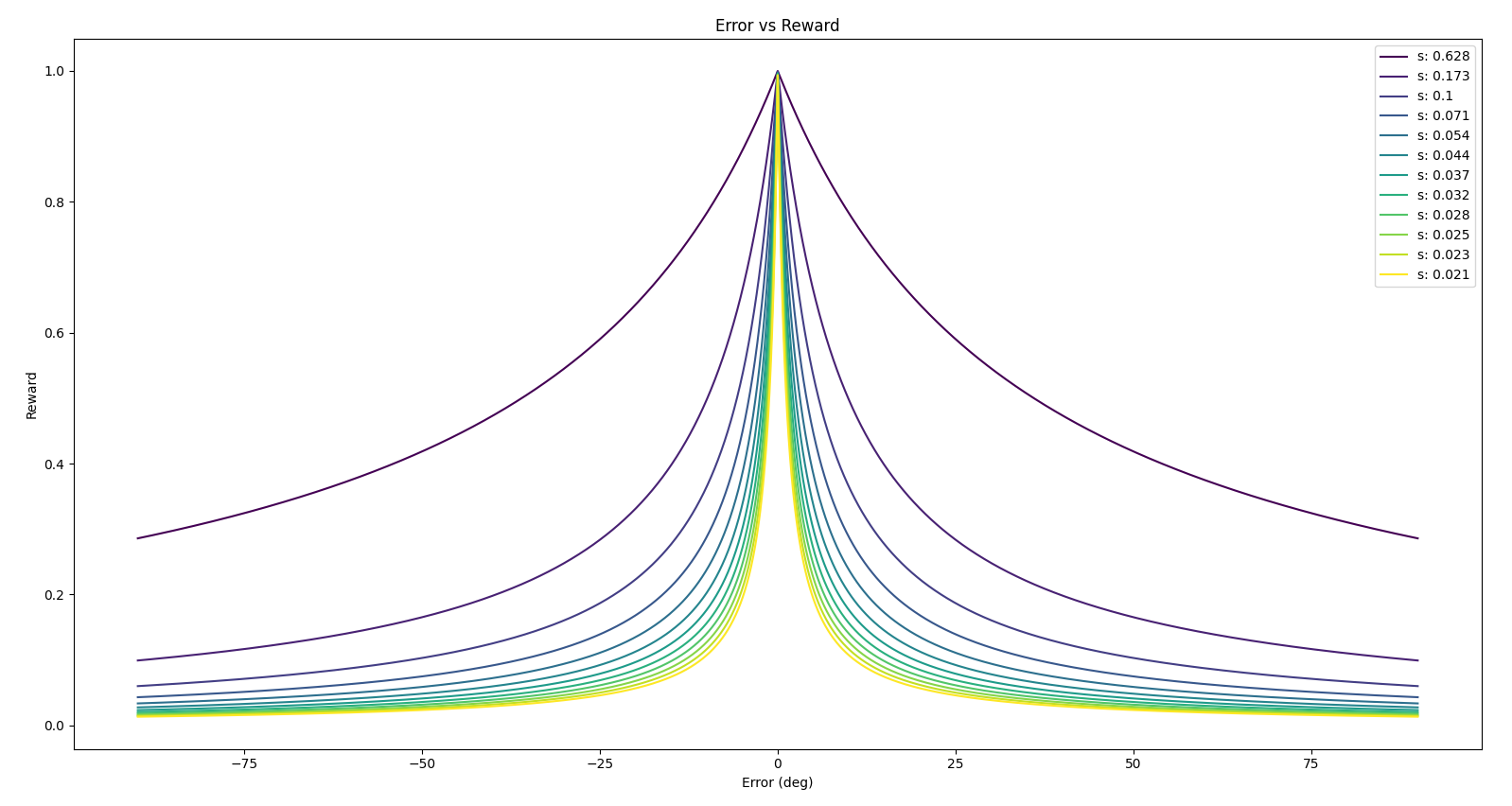}}
\caption{Change of reward with respect to error for different scaling factors.}
\label{error-rew-fig}
\end{figure*}

\section{Results}\label{result}
In this section, the results of our study are presented. In the initial phase, we trained a model for each maneuver type. Training time for different models is closely related to maneuver length and complexity. Initial trials with default hyper-parameters of the SAC algorithm produced unsatisfactory results, therefore we used hyper-parameter optimization and trained the models with optimized hyper-parameters later. Reference maneuver trajectories for different maneuvers along with the trajectories generated by the trained models are displayed in the Figures \ref{loop-normal}, \ref{immelmann-normal}, and \ref{barrel-roll-normal}. In the training process, we noted that longer maneuvers require more training time and are inclined to accommodate errors, therefore, exhaustive hyper-parameter optimization is applied for longer maneuvers such as loop. Also, we utilized two different termination conditions. In the first one, we terminated the episode if the agent trajectory diverged significantly from the target trajectory. To calculate divergence, we defined various equations for each maneuver, basically if the maneuver is mostly on the longitudinal axis, we terminated the episode if the gamma error increased above 30 degrees. If the maneuver is mostly in the lateral axis, we terminated the episode when the roll and yaw error increased above 60 degrees. These values are determined via experimentation. In the second termination condition, we only terminated the episode when the total maneuver time was reached. After trials with both termination conditions, we saw that there was no difference in training performance or time, and we continued with the second termination condition.

In Figures \ref{loop-handcrafted}, \ref{immelmann-handcrafted}, and \ref{barrel-roll-handcrafted}, results for the models that are trained with artificially generated trajectories are displayed. Like previous ones, RL models are capable of learning the desired maneuver when the provided trajectory is physically meaningful. Beyond the domain knowledge, we spend some time on tuning the parameters of these trajectories such as gamma increase rate, roll change rate, etc. In terms of convergence, both types of trajectories resulted in good learning performance. However, maneuvers generated with artificially created trajectories tend to be smoother due to their noise-free structure, which leads to the perception of AI pilots as less human-like. It is important to emphasize that domain knowledge is the most significant factor when designing a maneuver trajectory, therefore, it is not a trivial task.

\begin{figure*}
\centerline{\includegraphics[width=1\textwidth]{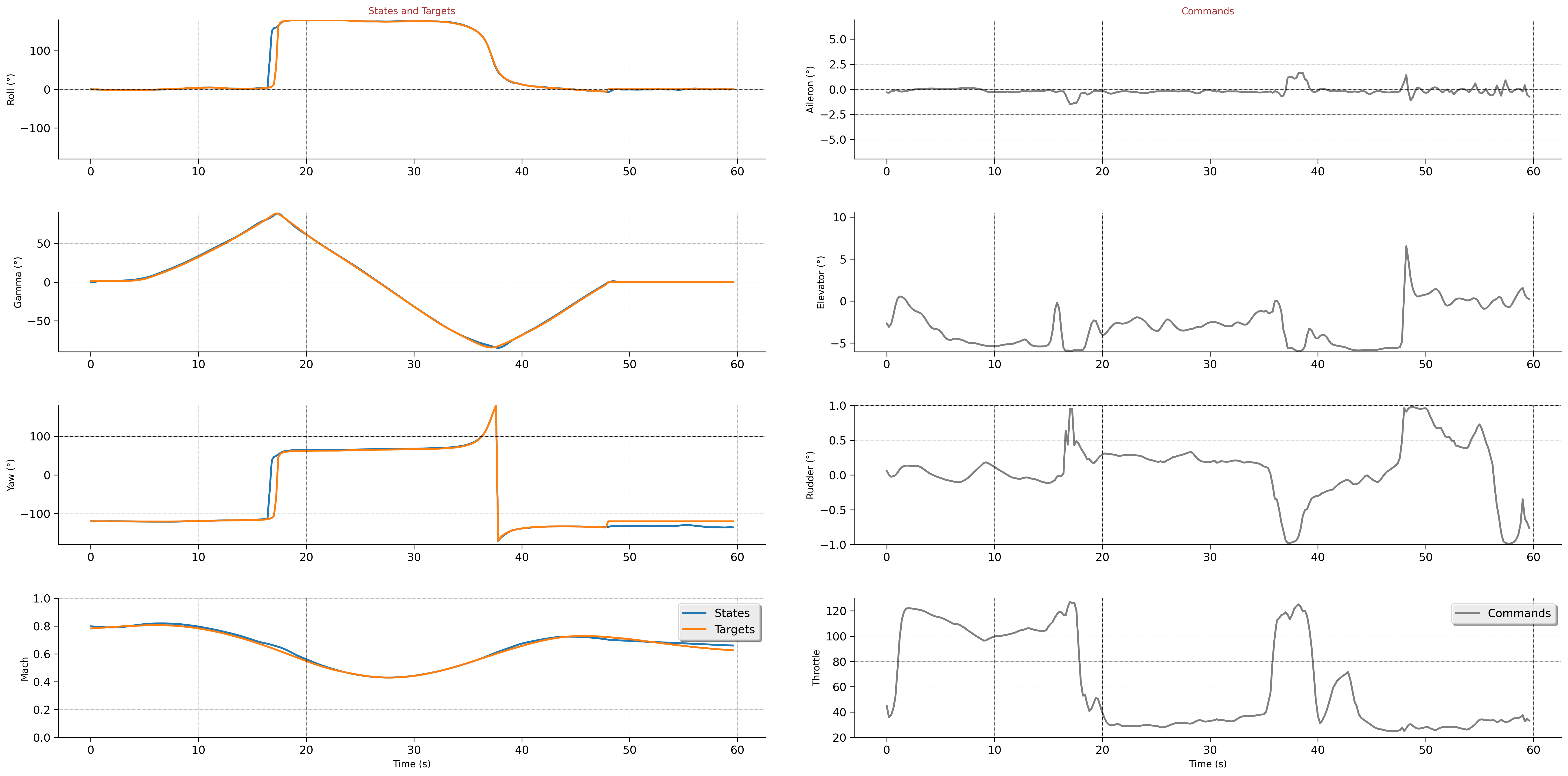}}
\caption{Target trajectory, resulting trajectory and actions applied by the RL model is presented for loop maneuver with pilot referenced target trajectory.}
\label{loop-normal}
\end{figure*}

\begin{figure*}
\centerline{\includegraphics[width=1\textwidth]{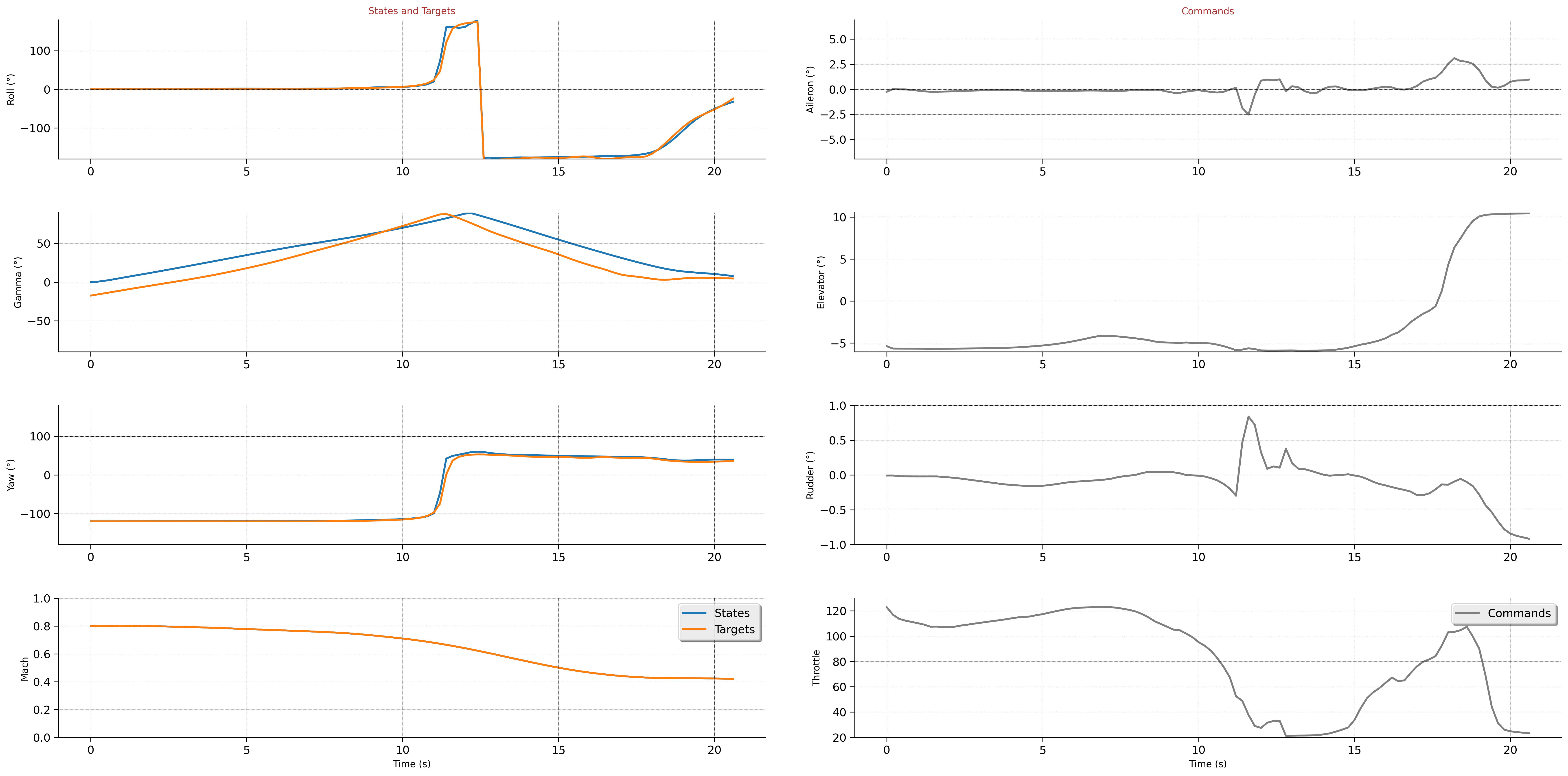}}
\caption{Target trajectory, resulting trajectory and actions applied by the RL model is presented for Immelmann maneuver with pilot referenced target trajectory.}
\label{immelmann-normal}
\end{figure*}

\begin{figure*}
\centerline{\includegraphics[width=1\textwidth]{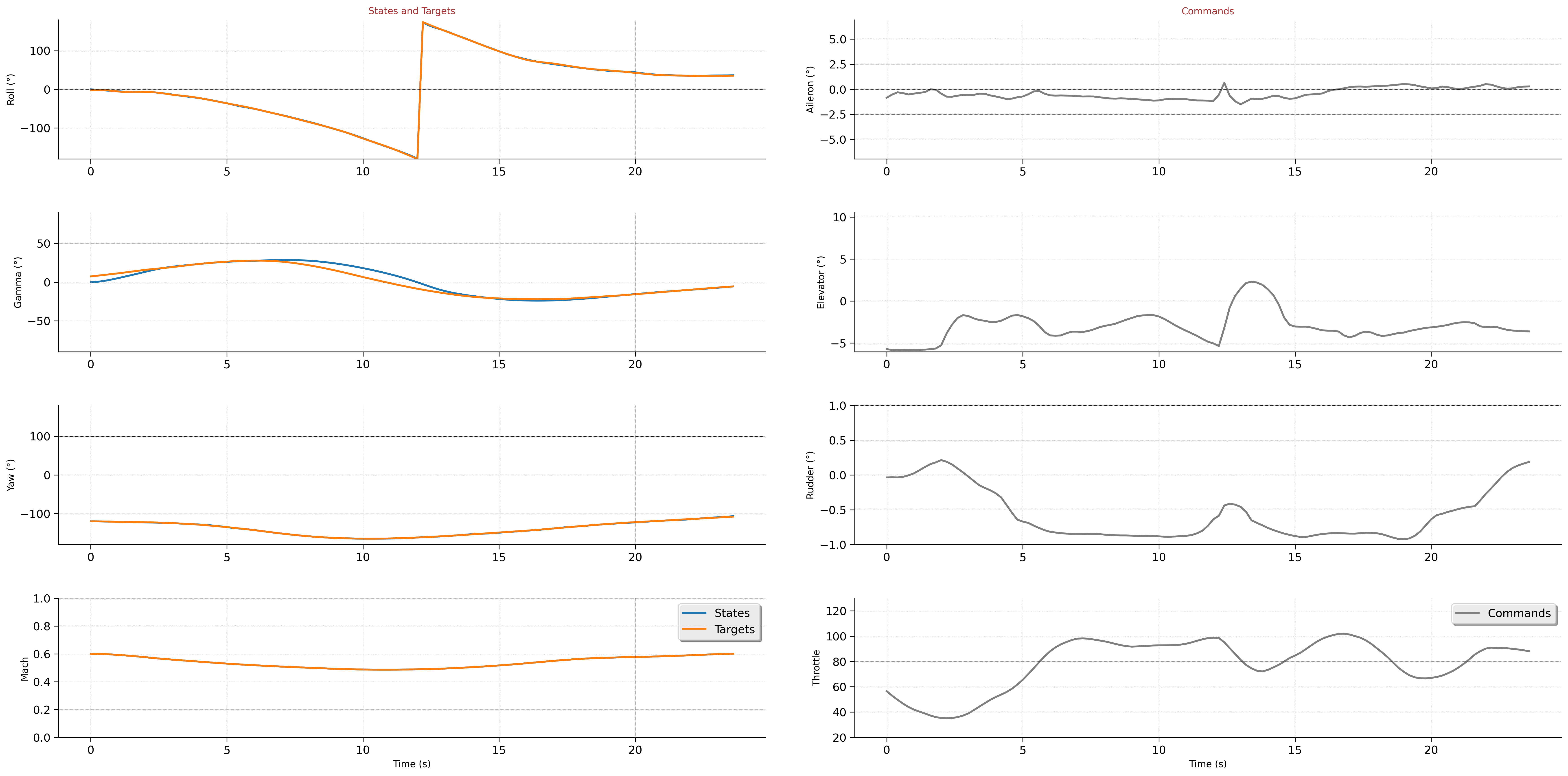}}
\caption{Target trajectory, resulting trajectory and actions applied by the RL model is presented for barrel roll maneuver with pilot referenced target trajectory.}
\label{barrel-roll-normal}
\end{figure*}

\begin{figure*}
\centerline{\includegraphics[width=1\textwidth]{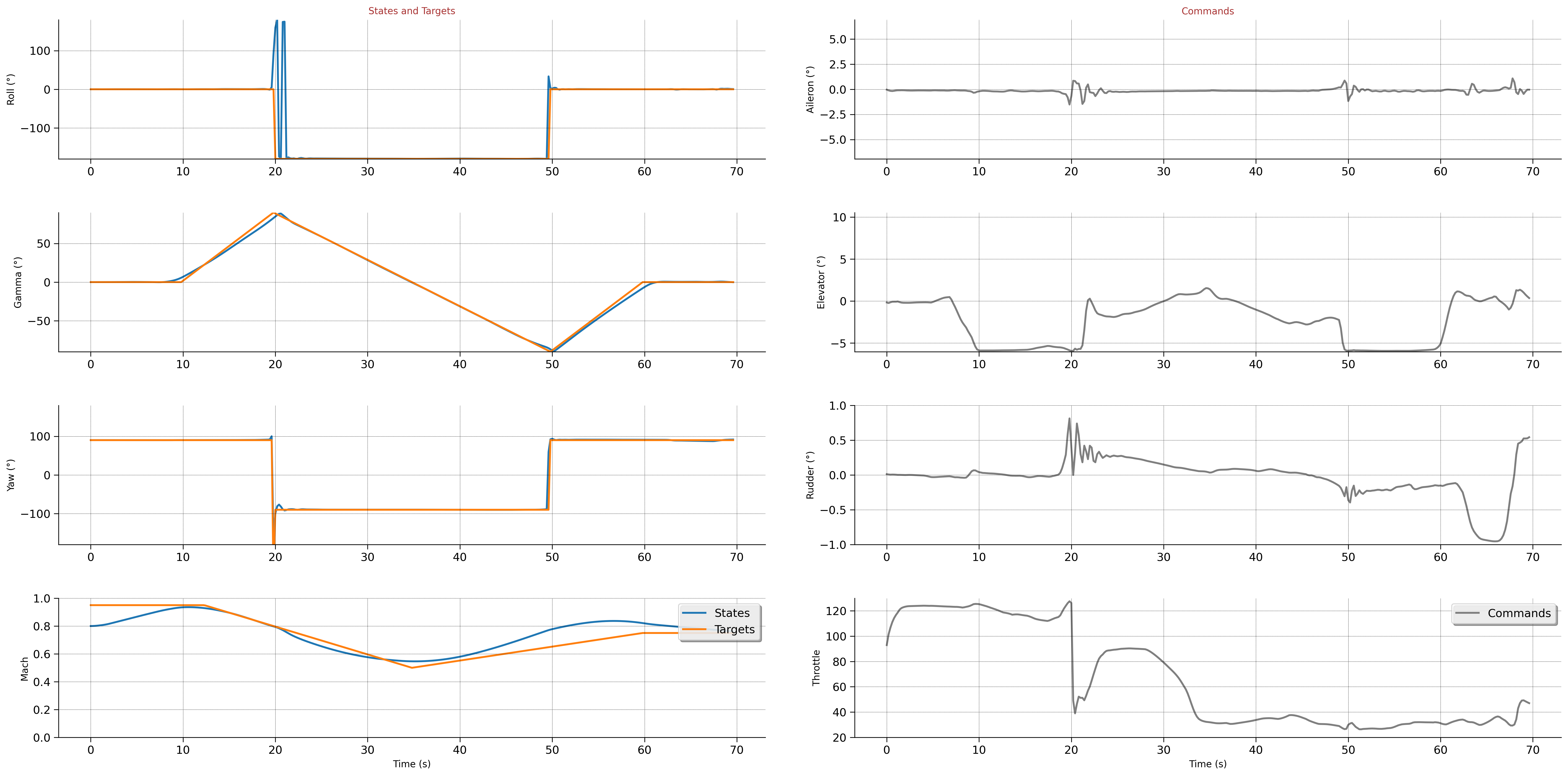}}
\caption{Target trajectory, resulting trajectory and actions applied by the RL model is presented for loop maneuver with handcrafted target trajectory.}
\label{loop-handcrafted}
\end{figure*}

\begin{figure*}
\centerline{\includegraphics[width=1\textwidth]{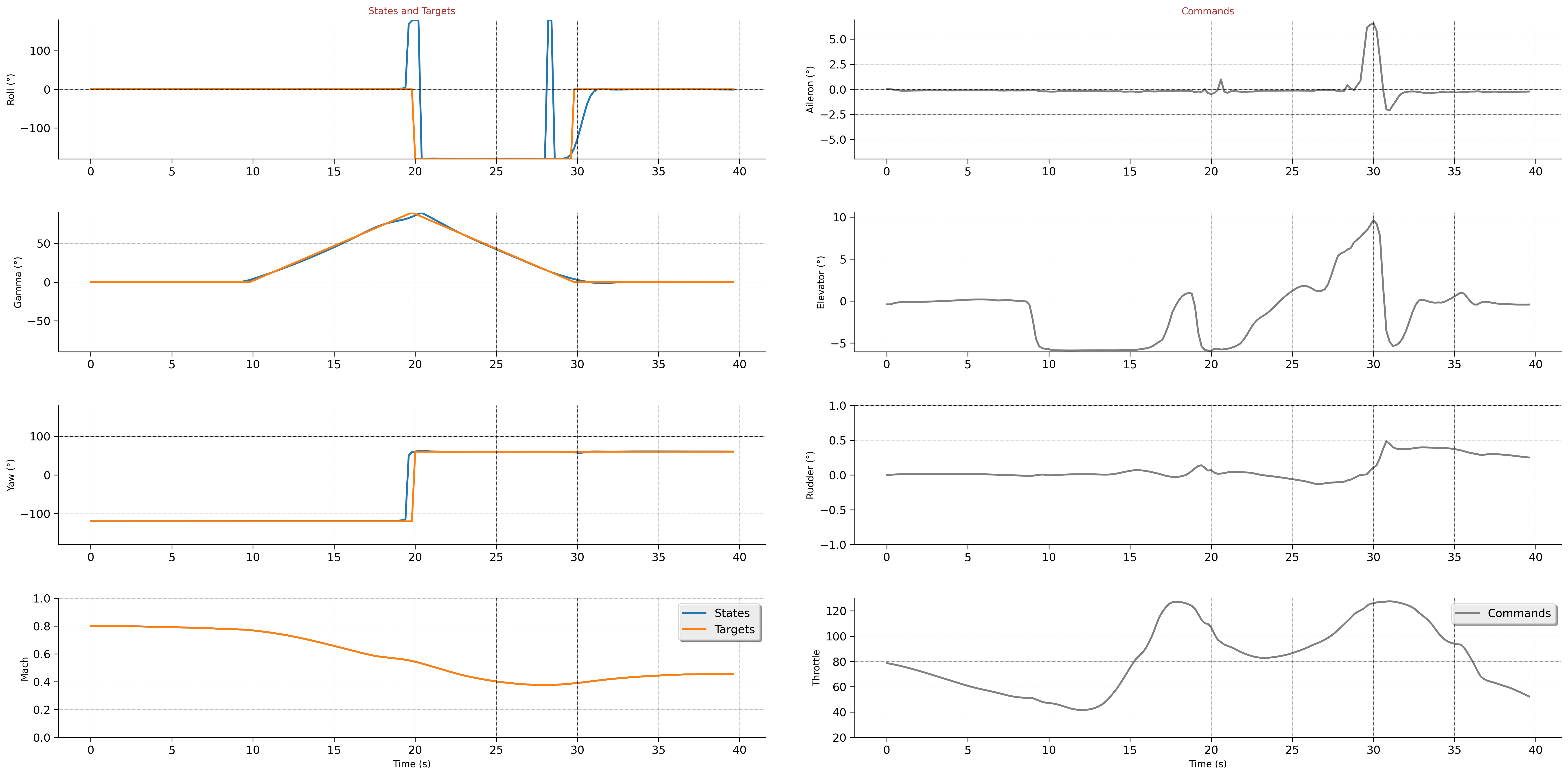}}
\caption{Target trajectory, resulting trajectory and actions applied by the RL model is presented for Immelmann maneuver with handcrafted target trajectory.}
\label{immelmann-handcrafted}
\end{figure*}

\begin{figure*}
\centerline{\includegraphics[width=1\textwidth]{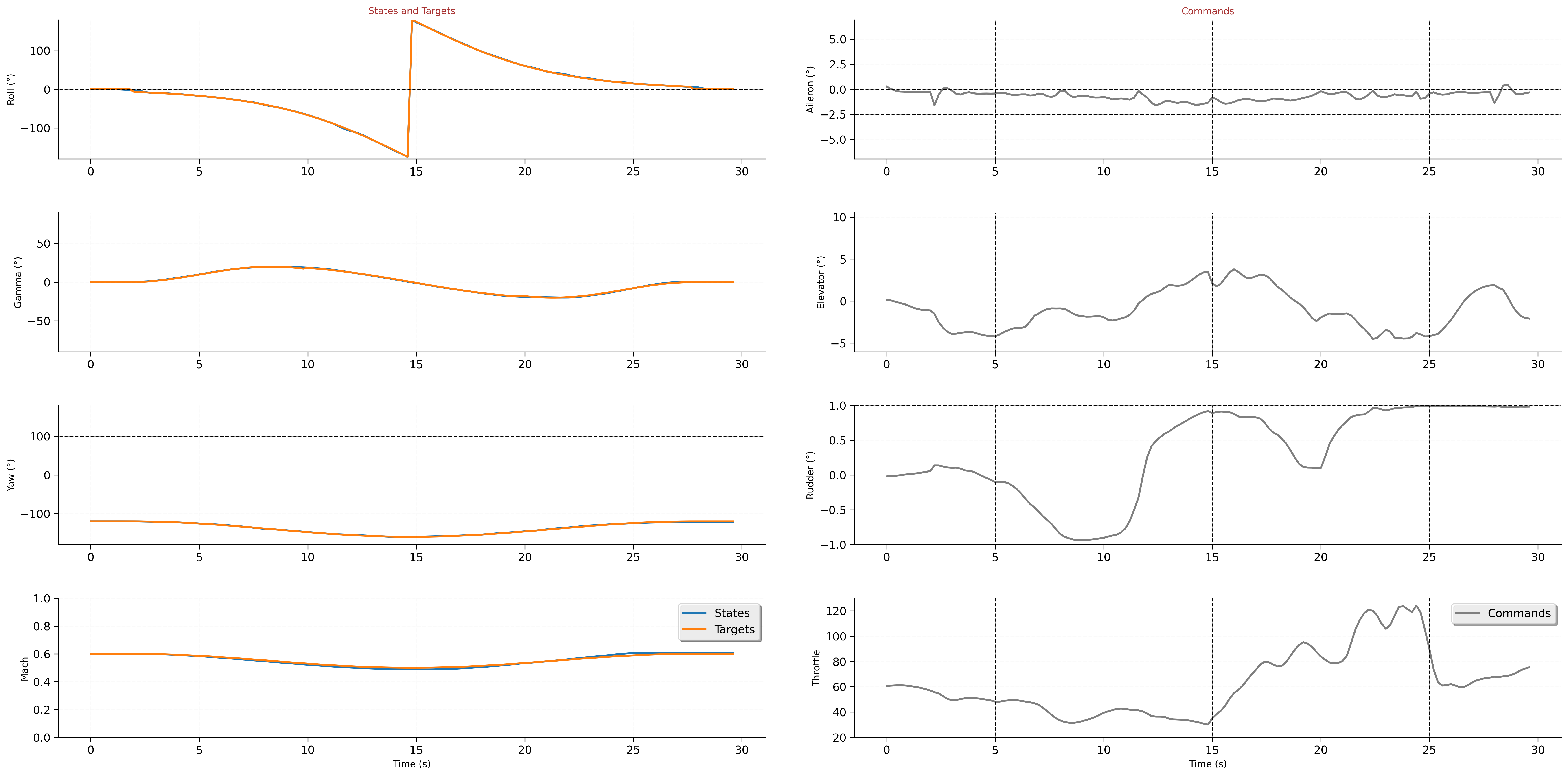}}
\caption{Target trajectory, resulting trajectory and actions applied by the RL model is presented for barrel roll maneuver with handcrafted target trajectory.}
\label{barrel-roll-handcrafted}
\end{figure*}

Beyond obtaining the desired trajectories, we also examined the performance of our models in a wide range of initial conditions. As explained in Section \ref{init}, training the models with relative yaw angle targets paved the way for independence from the initial yaw angle. Our models are able to execute a maneuver in different initial directions along with the different initial positions in terms of latitude and longitude. However, changing the initial altitude results in some distortions in the maneuver if the maneuver is laterally oriented, like a barrel roll, since the model did not see significant changes in altitude during training. To solve that problem, it is necessary to train the models by involving various initial altitude conditions and increasing the training budget, which can be done by increasing the training time and computational resources. In addition to the maneuvers described in this paper, we experimented with 27 custom trajectories created by pilots, and we observed similar learning performance for them. These experiments suggest that our approach can be used to learn any desired maneuver/trajectory as long as it is physically acceptable.

Another important result is the applicability of time scaling to a desired maneuver trajectory. As explained in the methodology part, we experimented with different time scaling factors to test whether our algorithm is able to execute maneuvers in different lengths or not. We have seen that trained models are able to execute the same maneuver at different lengths by using the same reference trajectory. However, training with a variable time scaling factor increased the training time significantly.  An important aspect of the time scaling feature is considering the feasibility of the resulting trajectory, which is quite significant when deciding on the time scaling factor. As an example, Hurjet aircraft has a pitch rate limit like every other aircraft and very short loop maneuvers (e.g. 15 seconds) are not physically possible. So, our study proves the applicability of this concept to acceptable physical constraints.

Stability of the trained models is an important concern, and we tested it by passing beyond the initial conditions used in the training phase. To generate naturally occurring initial conditions, we concatenated different maneuvers. The concatenation process created more physically acceptable initial conditions between maneuver changes. As an example, when we execute two loop maneuvers consecutively, the second loop maneuver starts with initial conditions that an actual pilot can observe in a real-life scenario. Under these conditions, our models were able to execute the desired maneuvers with similar quality. These findings suggest that trained models can be used to generate more complex trajectories and it is possible to cover the flight envelope if a sufficient number of models (maneuvers) are obtained.

Another significant finding of our work is that the quality of the results heavily depends on hyper-parameter optimization. Adjusting weight factors in the reward function is also important; however, we have seen that exhaustive hyper-parameter optimization can result in equally capable models even if weight factors are different. On the other hand, hyper-parameter optimization takes a significant amount of time (maybe up to one week). Therefore, we mostly worked on adjusting weight factors for different maneuvers to improve the results. 

Visualizations of the discussed maneuvers are available on \cite{youtube} as videos.

\section{Conclusion}\label{conclusion}
In this study, we aimed to generate models that can precisely execute aerobatic maneuvers on F16 and Hurjet aircraft, and we used a supervised learning-like approach to create the reward function for the SAC algorithm. We used pilot-generated trajectories and artificially created trajectories as references to our algorithm. By utilizing our proposed approach, we successfully obtained reinforcement learning models that can execute various aerobatic maneuvers. As an extension to our work, we introduced time scaling to our approach and proved the applicability of the concept by obtaining different-sized loop maneuvers by using the same reference trajectory. Our results suggest that the proposed approach can be used for any physically acceptable maneuver and generates stable and robust models. 

In light of our findings, it is possible to say that basic maneuvers can be used to construct more complex maneuvers, which is also discussed in the study \cite{nkemal2012}. In future work, we plan to obtain a basic maneuver set by using reinforcement learning and show the possibility of covering the flight envelope with these basic maneuvers.

\section*{Acknowledgement}
The project was realized as part of Hurjet advanced jet trainer platform's artificial intelligence concept projects within Turkish Aerospace. The test pilots were employees of the company. The simulation environment and flight dynamics model of the jet were provided by the company.

\bibliographystyle{ieeetr}
\bibliography{bibi}

\end{document}